\documentclass[letterpaper, 10 pt, journal, twoside]{IEEEtran}
\usepackage{graphicx}
\usepackage{amsmath}
\usepackage{xcolor} 
\usepackage{amssymb}
\usepackage{booktabs}
\usepackage{multirow}
\usepackage{balance}
\usepackage{makecell}
\usepackage{float}
\usepackage{amsmath,amsfonts}
\usepackage{array}
\usepackage[caption=false,font=normalsize,labelfont=sf,textfont=sf]{subfig}
\usepackage{textcomp}
\usepackage{stfloats}
\usepackage{url}
\usepackage{verbatim}
\usepackage{graphicx}
\usepackage{cite}
\usepackage[linesnumbered,ruled,vlined]{algorithm2e}
\SetKwInput{KwIn}{Input}
\SetKwInput{KwOut}{Output}
\SetAlgoNlRelativeSize{-1}  
\DontPrintSemicolon   

\begin{document}

\title{SCOPE: Skeleton Graph-Based Computation-Efficient Framework for Autonomous UAV Exploration}
\bstctlcite{BSTcontrol}
\author{Kai Li, Shengtao Zheng, Lingkun Xiu, Yuze Sheng, Xiao-Ping Zhang, $\textit{Fellow, IEEE}$, \\ Dongyue Huang, $\textit{Member, IEEE}$, Xinlei Chen, $\textit{Member, IEEE}$
\thanks{Manuscript received: October 15, 2025; Revised: January 13, 2026; Accepted: February 17, 2026.} 
\thanks{This paper was recommended for publication by Editor Aniket Bera upon evaluation of the Associate Editor and Reviewers' comments. This work was supported by the Natural Science Foundation of China under Grant 62371269, Shenzhen Science and Technology Program ZDCYKCX20250901094203005, Shenzhen Low-Altitude Airspace Strategic Program Portfolio (Grant No. Z25306110) and Meituan Academy of Robotics Shenzhen. $\textit{(Kai Li and Shengtao Zheng contributed equally to this work.)}$ $\textit{(Corresponding authors: Dongyue Huang; Xinlei Chen.)}$}
\thanks{Kai Li, Shengtao Zheng, Lingkun Xiu, Yuze Sheng, Xiao-Ping Zhang and Xinlei Chen are with the Shenzhen International Graduate School, Tsinghua University, Shenzhen 518055, China  (e-mail: \{likai24, st{\_}zheng24, xlk24, shengyz24\}@mails.tsinghua.edu.cn, xpzhang@ieee.org, chen.xinlei@sz.tsinghua.edu.cn).}%
\thanks{Dongyue Huang is with the School of Electrical and Electronic Engineering, Nanyang Technological University, Singapore 639956 (e-mail: dongyue.huang@ntu.edu.sg).}%
\thanks{Digital Object Identifier (DOI): see top of this page.}
}


\markboth{IEEE ROBOTICS AND AUTOMATION LETTERS. PREPRINT VERSION. ACCEPTED FEBRUARY, 2026}
{Li \MakeLowercase{\textit{et al.}}: SCOPE: Skeleton Graph-Based Computation-Efficient Framework for Autonomous UAV Exploration}


\maketitle

\begin{abstract}
Autonomous exploration in unknown environments is key for mobile robots, helping them perceive, map, and make decisions in complex areas. However, current methods often rely on frequent global optimization, suffering from high computational latency and trajectory oscillation, especially on resource-constrained edge devices. To address these limitations, we propose SCOPE, a novel framework that incrementally constructs a real-time skeletal graph and introduces Implicit Unknown Region Analysis for efficient spatial reasoning. The planning layer adopts a hierarchical on-demand strategy: the Proximal Planner generates smooth, high-frequency local trajectories, while the Region-Sequence Planner is activated only when necessary to optimize global visitation order. Comparative evaluations in simulation demonstrate that SCOPE achieves competitive exploration performance comparable to state-of-the-art global planners, while reducing computational cost by an average of 86.9\%. Real-world experiments further validate the system's robustness and low latency in practical scenarios. 
\end{abstract}


\section{Introduction}
Autonomous Uncrewed Aerial Vehicles (UAV) exploration technology plays a crucial role in intelligent uncrewed systems \cite{chen2024ddl}, offering vast potential for applications in disaster response and victim localization \cite{khan2022emerging}, structural inspection of large-scale infrastructures \cite{altshuler2017optimal}, and 3D reconstruction of complex environments \cite{zhang2024autofusion, Li2025}. Among the existing approaches, sampling-based strategies \cite{bircher2016receding} rely on probabilistic search techniques to ensure environmental coverage, frontier-based approaches \cite{deng2020robotic} drive exploration by identifying frontiers between known and unknown regions, and hybrid search approaches \cite{10816079} adopt hybrid frameworks that combine multiple strategies to enhance overall exploration efficiency. However, as environments grow in scale and complexity, the computational overhead of exploration planning becomes a critical bottleneck. On resource-constrained UAVs \cite{ding2025hawkeye}, these algorithms must coexist with not only safety-critical tasks like perception and avoidance but also intelligence-heavy missions such as real-time target recognition and semantic analysis \cite{gu2025mr,zha2025dimm}. Reducing computational costs is therefore essential for providing the necessary `headroom' to ensure both system robustness and multi-tasking capability.

Despite this critical requirement, many advanced methods \cite{10816079,zhou2021fuel,zhao2023autonomous,bu2025rush} rely on frequent global Asymmetric Traveling Salesman Problem (ATSP) solvers, which presents three main challenges. 
First, the exponential computational complexity of ATSP imposes a heavy overhead on resource-constrained edge devices, limiting re-planning frequency. Second, exploration is a Partially Observable Markov Decision Process (POMDP) where frontiers evolve continuously. This dynamic nature triggers topological oscillations—where minor observations cause drastic flips in the global tour, leading to unstable flight intentions. Third, in large-scale maps, global cost functions are often dominated by noisy heuristic estimates of unknown areas. This uncertainty can drown out the precise costs of the known local region in a global sum-objective, while the ``plan-all, execute-one'' nature of ATSP further wastes computation and leads to redundant coverage by deferring small residual regions to the end of the ATSP tour.

\begin{figure*}[t]
    \centering
    \includegraphics[width=16.5cm]{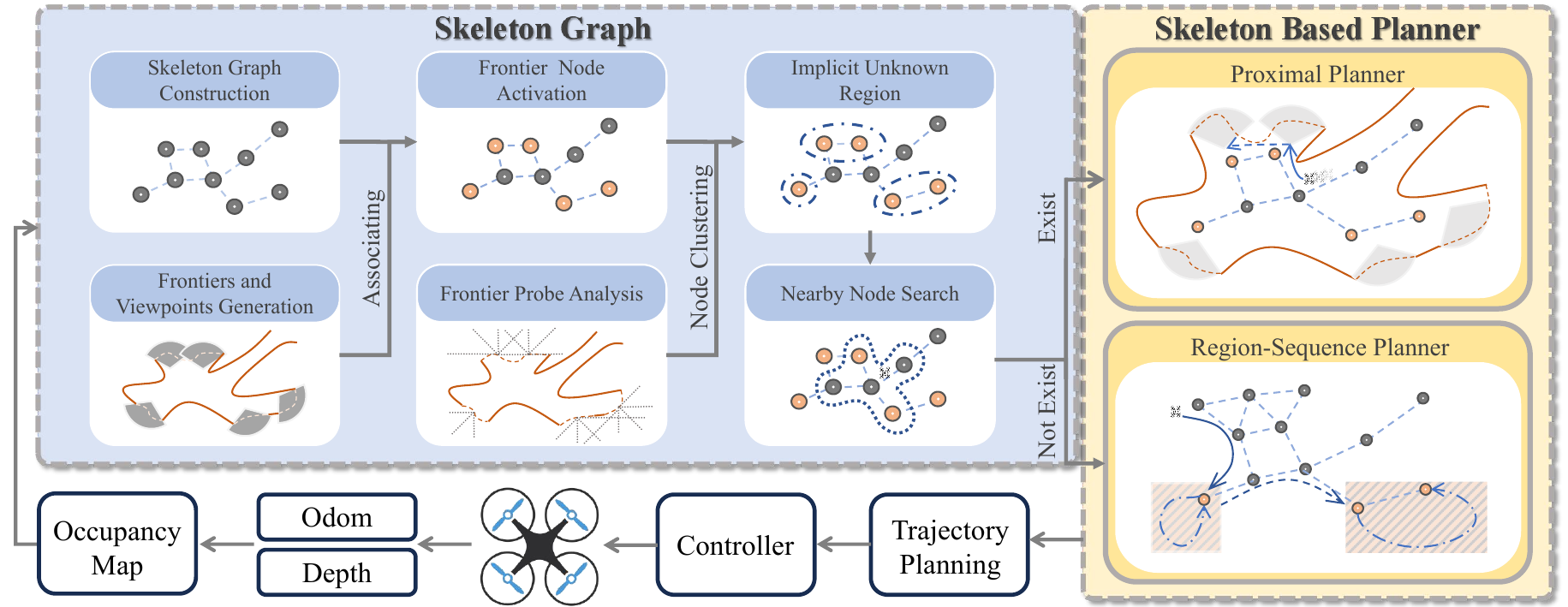}
\caption{System overview of the proposed SCOPE framework. The architecture integrates two subsystems: (1) Skeleton Graph Maintenance \& Implicit Region Generation (Blue Blocks): The system constructs a skeleton graph and activates nodes associated with frontiers. Using geometric ray probes for depth estimation, activated nodes are clustered into Implicit Unknown Regions---abstract volumetric representations where clustered nodes likely lead to the same contiguous unknown area. (2) Skeleton-Based Hierarchical Planning (Yellow Blocks): The UAV searches for a proximal target in its vicinity. If found (Exist), the Proximal Planner executes rapid local exploration; otherwise (Not Exist), the Region-Sequence Planner optimizes a global sequence over implicit sub-regions to guide the UAV to the next strategic area.}
    \label{framework}
\end{figure*}

To address these challenges, we present SCOPE, a skeleton graph-based framework that adopts an on-demand planning strategy to minimize computational overhead. 
Our framework incrementally maintains a geometric topology-based skeletal graph to capture the environment's topology. 
Crucially, we introduce Implicit Unknown Region Analysis using geometric probes, which allows the system to efficiently reason about the connectivity of unexplored space without explicit voxel-level processing. 
Building upon this, the Proximal Planner generates smooth, high-frequency trajectories for local coverage, unaffected by topological oscillations, while the Region-Sequence Planner is activated only when necessary (e.g., when local targets are exhausted) to provide global guidance. 
This hierarchical design achieves a balance between global optimality and local agility with minimal computational cost.

We compared the proposed method with four baseline approaches in various simulation environments. 
Experimental results show that SCOPE maintains competitive exploration efficiency comparable to state-of-the-art global planners, while achieving the lowest computational cost across all scenarios.
Additionally, the method was deployed on a real-world UAV to validate its effectiveness. 
The contributions of this work are summarized as follows:

\begin{itemize}
    \item We propose an implicit space decomposition method that incrementally constructs a geometric topology-based skeletal graph and performs unknown space analysis via geometric probes.
    
    \item We design an on-demand hierarchical planning framework that synergizes a high-frequency Proximal Planner with a low-frequency Region-Sequence Planner to minimize computational overhead.
    
    \item We conduct extensive simulations and real-world UAV experiments to validate the efficacy of the proposed framework.

\end{itemize}

\section{Related Work}
\subsection{Autonomous Exploration}
Autonomous exploration methodologies are primarily categorized into frontier-based, sampling-based, and hybrid approaches. Frontier-based methods, pioneered by Yamauchi \cite{yamauchi1997frontier, yamauchi1998frontier, thrun2003learning}, prioritize boundaries between known and unknown regions. Subsequent works have enhanced this by integrating motion constraints \cite{gao2018improved} and trajectory optimization \cite{deng2020robotic} to improve execution quality. In contrast, sampling-based strategies \cite{bircher2016receding} utilize Rapidly-exploring Random Trees (RRT) to generate Next-Best-Views (NBV) based on information gain. However, these methods typically rely on local-greedy heuristics and lack global context, often resulting in redundant backtracking and reduced efficiency.

Hybrid frameworks aim to balance exploration efficiency and completeness by combining multiple strategies. FUEL \cite{zhou2021fuel} introduces a hierarchical planning mechanism to alleviate exhaustive local searches, though it overlooks the dynamic nature of the exploration process. FAEP \cite{zhao2023autonomous} adopts a spiral filling strategy but is constrained by low flight speeds. While Yang et al. \cite{yang2021graph} utilize sparse topological maps, their greedy global policy limits performance in complex environments. More recently, RACER \cite{zhou2023racer} and FALCON \cite{10816079} decompose the unknown space before planning. However, the frequent solving of the ATSP or Sequential Ordering Problem (SOP)---coupled with the ``plan-all, execute-one'' paradigm---imposes a heavy computational burden.

\subsection{Skeleton Graph}

In robotic navigation, skeleton graphs are utilized to reduce the search space for long-horizon path planning \cite{beom1993path}. Prior research primarily focused on efficient skeleton extraction and robust environment representation \cite{chen2022fast}, \cite{rezanejad2015robust}, laying the foundation for their integration into exploration strategies. 

While several methods employ general topological graphs to guide exploration~\cite{wang2020efficient}, \cite{10801613}, many of them rely on probabilistic sampling \cite{respall2021fast, xu2021autonomous}. In contrast, skeletons provide a more faithful representation by explicitly capturing the medial axis of the free space. This geometric nature yields a compact and stable structure, avoiding the randomness and redundancy inherent in stochastic roadmaps, making them particularly suitable for guiding exploration in complex 3D environments. 

Their primary advantage lies in facilitating efficient global planning by capturing the essential connectivity of an unknown space, while also improving robustness against local noise and reducing redundant paths through compact free-space abstraction. However, skeletons alone cannot directly reflect the distribution of unexplored regions, necessitating the integration of complementary modules for unknown space analysis.

\section{Methods}

This letter proposes SCOPE, a computation-efficient exploration framework designed to address the challenges of high dimensionality, flight instability, and redundant coverage. As illustrated in Fig.~\ref{framework}, the system operates through three synergistic modules. First, to address the computational complexity of high-dimensional space, the preprocessing stage incrementally constructs a geometric topology-based skeletal graph combined with Implicit Unknown Region Analysis. By utilizing geometric probes, this stage effectively decomposes and assesses unknown space to provide a compact representation for subsequent planning. Second, to mitigate flight instability caused by frequent global re-planning, the proximal planner operates at a high frequency, generating smooth, local trajectories by prioritizing spatially continuous frontiers. Finally, to resolve the ``plan-all, execute-one'' inefficiency and escape local minima, the region-sequence planner is activated in an on-demand manner. This module leverages the skeletal graph to solve a symmetric Traveling Salesman Problem (TSP) only when local exploration targets are exhausted, providing global guidance with minimal computational overhead.

\subsection{Skeleton Graph} \label{prepro}

In the proposed framework, the skeleton graph of free space serves as the core environment representation, capturing the 3D structure and obstacle distribution. The graph is incrementally constructed from a voxel grid of the Euclidean Signed Distance Field (ESDF), which is uniformly downsampled by a fixed factor $N$ to filter sensor noise and reduce computational overhead. Key topological nodes are identified as local maxima whose ESDF values exceed those of their 26 discrete 3D neighbors, subject to a minimum inter-node distance threshold $d_{\rm{thr}}$. Consequently, due to the combined effects of these discretization and sparsity constraints, nodes may not strictly align with the continuous geometric medial axis, but this approximation significantly reduces redundancy. Furthermore, graph updates are strictly confined within the axis-aligned bounding box (AABB) of the map update range to minimize overhead. Edges are then established between nodes if three criteria are met: 1) the inter-node distance is within a connection threshold $d_{\rm{conn}}$; 2) the direct connection is collision-free; and 3) the angle formed with existing connected edges exceeds a specific angular threshold $\theta_{\rm{thr}}$.

After graph construction, frontiers are assigned to the nearest skeletal nodes via Euclidean-distance-based search with ray-casting visibility checks. A skeleton node is marked as activated if it is assigned at least one frontier point, and deactivated otherwise.

\begin{figure}[t]
    \centering
    \includegraphics[width=8.0cm]{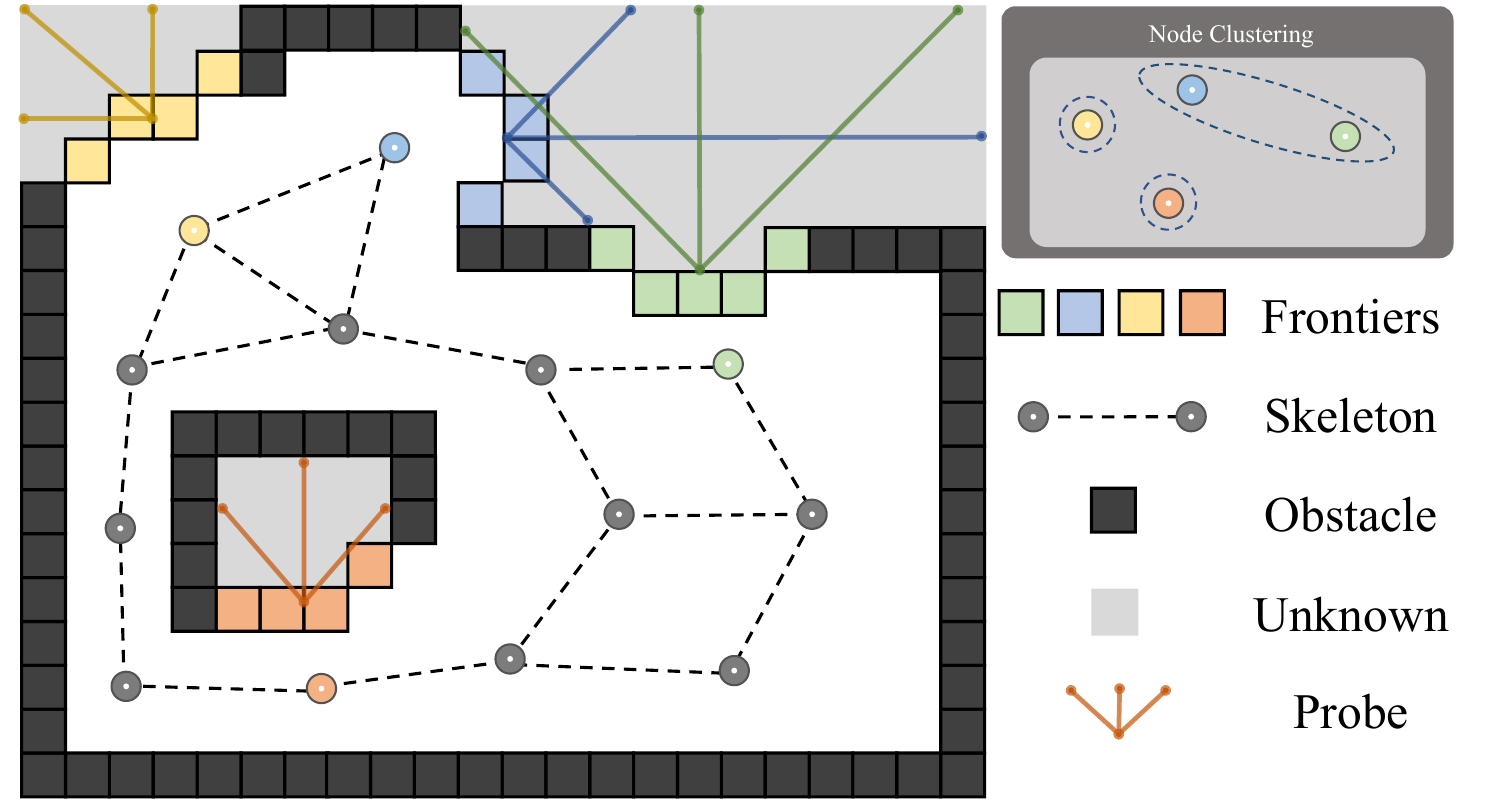}
    \caption{Region analysis: probes are cast from the frontiers associated with activated nodes into unknown space, and nodes are clustered based on their intersection patterns and spatial proximity.}
    \label{region_analysis}
\end{figure}

The distance between two points \(P_1\) and \(P_2\) in the skeletal graph and the hop distance between these points are two critical metrics used in path planning. The skeletal distance, \(d_{\text{skel}}(P_1, P_2)\), is defined as the sum of the Euclidean distances along the shortest path connecting the two points in the graph:

\begin{equation}
   d_{\text{skel}}(P_1, P_2) = \sum_{i=1}^{n-1} \| P_i - P_{i+1} \|, 
\end{equation}
where \(P_1, P_2, \dots, P_n\) represents the sequence of nodes in the shortest path between \(P_1\) and \(P_2\), and \(\| P_i - P_{i+1} \|\) is the Euclidean distance between consecutive nodes in the path.

The hop distance, \(h_{\text{skel}}(P_1, P_2)\), is defined as the number of edges in the shortest path between \(P_1\) and \(P_2\):
\begin{equation}
    h_{\text{skel}}(P_1, P_2) = \min \{ |P_1, P_2| \mid P_1 \xrightarrow{\text{path}} P_2 \},
\end{equation}
where \(|P_1, P_2|\) denotes the number of edges in that path.

\subsection{Implicit Unknown Region Analysis}

After constructing the skeletal graph, the system performs Implicit Unknown Region Analysis (Algorithm~\ref{alg:region_analysis}, Fig~\ref{region_analysis}) to enhance the global awareness of the exploration. The goal is to partition the skeletal nodes into structured regions based on their sensory coverage of the unknown space.

\subsubsection{Geometric Probes}
To quantify the potential volume of the unknown space, the system employs a stratified probing strategy (Line~\ref{line:probe}). The local map is analyzed in vertical layers, and eight-directional 2D geometric probes are uniformly cast from the centroid of each frontier cluster. Specifically, these probes are designed to estimate the potential depth of the unknown space. For each ray, the penetration depth is measured starting from the position where the ray first resides inside the unknown space within a distance threshold $d_{\text{init}}$. If the ray fails to enter the unknown space within this threshold or encounters an obstacle, it is considered invalid for depth calculation. For valid rays, the propagation normally terminates when it either re-enters known space or reaches the boundary of the map. Furthermore, if the ray propagation exceeds a maximum sensing range $D_{\text{max}}$ before reaching such limits, it is terminated and the depth is strictly clamped to $D_{\text{max}}$. By measuring the distance between these entry and termination points, we capture the geometric distribution of the unknown volume in all directions.

\subsubsection{Region Clustering}
Subsequently, the system analyzes the spatial correlation between nodes to form regions (Lines~\ref{line:criteria}-\ref{line:union_end}). Unlike heuristic grouping, we define two nodes $v_i$ and $v_j$ as belonging to the same region if and only if they satisfy two rigorous criteria:
\begin{enumerate}
    \item \textbf{Spatially Close:} The Euclidean distance between them is within a proximity threshold, i.e., $d(v_i, v_j) < D_{\text{prox}}$, ensuring they belong to the same local neighborhood.
    \item \textbf{Sufficient Intersection:} Their probe rays exhibit a high degree of overlap, where the number of intersecting rays exceeds a validation threshold $N_{\text{thr}}$, indicating they observe the same continuous unknown volume.
\end{enumerate}

The system uses the Union-Find method to efficiently cluster all nodes satisfying these conditions into connected regions. Isolated nodes that do not form strong connections are treated as independent regions. For a region where the number of skeletal nodes exceeds a predefined threshold $T_{\rm{size}}$, K-means clustering is further applied to subdivide it based on centroid distribution (Line~\ref{line:kmeans}), ensuring spatial compactness.

\subsubsection{Isolation Assessment}
Once partitioned, we evaluate the exploration priority of each region (Line~\ref{line:iso}). To address the discrepancy between region size and accessibility, we define an Isolation Score $S_{\rm{iso}}$. $S_{\rm{iso}}$ is formulated strictly based on topological connectivity and depth:
\begin{equation}
   S_{\text{iso}} = \frac{1}{1 + \alpha N_{\text{ext}} + \beta \| D_{\text{avg}} \|_2},
   \label{S_iso}
\end{equation}
where \( N_{\text{ext}} \) is the number of intersection connections with other regions, \( D_{\text{avg}} \) is the average unknown depth of probes within the region, and \( \alpha \) and \( \beta \) are weight parameters. This metric assesses the region's independence and exploration value.

\begin{algorithm}[t]
\caption{Implicit Unknown Region Analysis}
\label{alg:region_analysis}

\KwIn{Activated nodes $\mathcal{V}_{act}$, Parameters $D_{\text{max}}, D_{\text{prox}}, N_{\text{thr}}$}
\KwOut{Clustered regions $\mathcal{R}$, Isolation scores $\{S_{iso}\}$}
\vspace{0.5em}

$D_{v,k} \gets \mathrm{ProbeDepth}(\mathcal{V}_{act}, D_{\text{max}})$\; \label{line:probe}
$C_{ij} \gets \mathrm{CountIntersections}(\mathcal{V}_{act})$\;

\ForEach{$(v_i, v_j) \in \mathcal{V}_{act}$ \textit{with} $C_{ij} \geq N_{\mathrm{thr}}$ \textit{and} $d_{\mathrm{skel}}(v_i, v_j) < D_{\mathrm{prox}}$}{ \label{line:criteria}
  $\mathrm{StrongConn} \gets \mathrm{StrongConn} \cup \{(v_i, v_j)\}$\;
}

\ForEach{$v \in \mathcal{V}_{act}$}{ \label{line:union_start}
  $\mathrm{parent}(v) \gets v$\;
}
\ForEach{$(v_i, v_j) \in \mathrm{StrongConn}$}{
  $\mathrm{Union}(v_i, v_j)$\;
}
\ForEach{$v \in \mathcal{V}_{act}$}{
  $r \gets \mathrm{FindRoot}(v)$\;
  $\mathrm{Assign}(v, r)$\; \label{line:union_end}
}

\ForEach{$R \in \mathcal{R}$}{ \label{line:kmeans}
  \If{$|R| > T_{\mathrm{size}}$}{
    $\mathrm{KMeans}(R)$\;
  }
}

\ForEach{$R \in \mathcal{R}$}{ \label{line:iso}
  $S_{\mathrm{iso}}(R) \gets $ Equation (\ref{S_iso})
}

\KwRet{$\mathcal{R}, \{S_{\mathrm{iso}}(R)\}$}\;

\end{algorithm}

\subsection{Proximal Planner}
The proximal planning module is responsible for efficiently generating viewpoint sequences in the vicinity of the UAV, enabling rapid and smooth exploration. The process begins by identifying $k$ neighboring skeletal graph nodes within the sensing range as starting points \cite{10801613}. From these nodes, the planner searches for active nodes with a hop distance $h_{\text{skel}}(P_1, P_2) < 3$. For each candidate node found, the time cost to reach its optimal viewpoint (defined as the pose providing the maximum information gain) from the current state $(P_c, \mathbf{v}_c)$ is computed. The candidate node associated with the minimum cost is then prioritized for expansion. This position-related cost $C_{\text{pos}}$ is defined as:

\begin{equation}
    C_{\text{pos}} = \frac{L}{v_{\max}} + t_{\text{init}} + \sum_{i} t_{\text{vel},i} + \frac{\Delta z}{v_{z,\max}},
\end{equation}
where \(L\) is the path length from the current position to the candidate viewpoint, \(v_{\max}\) is the maximum velocity, \(\Delta z\) is the total altitude change, and \(v_{z,\max}\) is the maximum vertical velocity. \(t_{\text{init}}\) is the time required to adjust the current velocity \(\mathbf{v}_1\) to the direction of the first segment, computed as:

\begin{equation}
   t_{\text{init}} = \frac{(v_{\max} - |v_c|)^2}{2 v_{\max} a_{\max}} + 
\begin{cases} 
2 \frac{|v_c|}{a_{\max}}, & v_c < 0 \\
0, & v_c \geq 0 
\end{cases} 
\end{equation}
where \(v_c = \mathbf{v}_1 \cdot \mathbf{d}_0\), \(\mathbf{d}_0\) is the unit vector of the first segment, and \(a_{\max}\) is the maximum acceleration. \(\sum_{i} t_{\text{vel},i}\) denotes the sum of time costs for velocity direction changes at each turning point along the sequence, calculated similarly to \(t_{\text{init}}\).

To avoid redundant revisits to already explored areas, the planner further prioritizes nodes in isolated regions---identified when the isolation score \(S_{\text{iso}}\) exceeds a predefined threshold---ensuring timely coverage and preventing large-scale backtracking in later stages of exploration.

After selecting the primary target, neighboring nodes within $h_{\text{skel}} \leq 1$ are identified as secondary targets. Rather than being visited immediately, these targets serve to optimize the primary viewpoint's yaw angle. By aligning the heading with their spatial distribution, the planner anticipates future exploration directions, ensuring visibility continuity \cite{zhou2021fuel} and preventing abrupt rotations. This refinement significantly enhances trajectory smoothness while maintaining efficiency.

\begin{figure}[t]
    \centering
    \includegraphics[width=8.0cm]{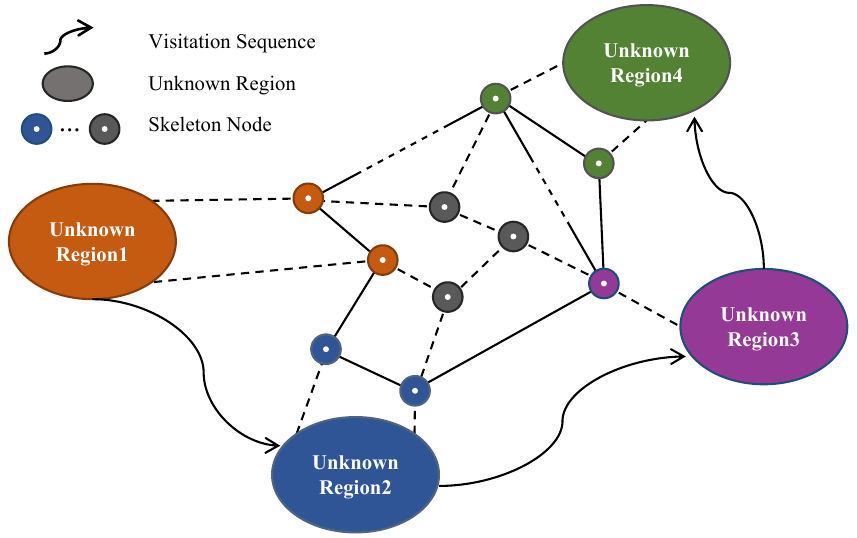}
    \caption{A schematic diagram of the region-sequence planner. During region-guided planning, it assumes exploration proceeds only through currently activated nodes into unknown regions, while inter-regional costs are computed on the skeletal graph.}
    \label{global_planner}
\end{figure}

\subsection{Region-Sequence Planner}

Region-sequence planning defines a high-level exploration order. Existing frontier-based planners \cite{zhou2021fuel} often neglect dynamic environmental changes, while partition-based methods \cite{10816079, zhou2023racer} suffer from high dimensionality and sensitivity to boundary definitions. To overcome these issues, we forego direct spatial modeling and instead cluster activated nodes (Section III-A) to implicitly represent target subregions. This approach significantly reduces optimization complexity and improves robustness, as shown in Fig.~\ref{global_planner}.

\begin{algorithm}[t]
\caption{Region-Sequence Planner}
\label{alg:region_sequence}
\KwIn{Clustered regions $\{V(A_1), \dots, V(A_n)\}$, skeletal graph $\mathrm{G_{skel}}$, current node $v_{\mathrm{cur}}$}
\KwOut{Next region $A^*$, target node $v^*$}
\vspace{0.5em}

$C_{\mathrm{tsp}} \gets \mathrm{InitMatrix}(n,n)$\; \label{line:matrix_start}

\For{$i \gets 1$ \KwTo $n$}{
  \For{$j \gets 1$ \KwTo $n$}{
    \eIf{$i = j$}{
      $C_{ij} \gets 0$\;
    }{
      $C_{ij} \gets \mathrm{MinSkelDist}(V(A_i), V(A_j), \mathrm{G_{skel}})$\;
    }
  }
}\label{line:matrix_end}

$\pi \gets \mathrm{LinKernighanHeuristic}(C_{\mathrm{tsp}})$\; \label{line:tsp}

$A^* \gets \pi(1)$\; \label{line:region_select}

$d_{\min} \gets \infty,\; v^* \gets \mathrm{None}$\;

\ForEach{$v \in V(A^*)$}{
  $d \gets d_{\mathrm{skel}}(v_{\mathrm{cur}}, v, \mathrm{G_{skel}})$\;
  \If{$d < d_{\min}$}{
    $d_{\min} \gets d,\; v^* \gets v$\;
  }
}\label{line:node_select}

\KwRet{$(A^*, v^*)$}\;

\end{algorithm}

Upon exhaustion of local targets, the region-sequence planner optimizes the global order by solving a TSP on clustered nodes (Line~\ref{line:tsp}). The resulting target directs the proximal planner toward the next strategic subregion, ensuring efficient global coverage (Lines~\ref{line:region_select}-\ref{line:node_select}).

The inter-region cost matrix \( C_{\text{tsp}} \in \mathbb{R}^{n \times n} \) is defined as (Lines~\ref{line:matrix_start}-\ref{line:matrix_end}):

\[
C_{\text{tsp}} =
\begin{bmatrix}
0 & C_{1,2} & \cdots & C_{1,n} \\
C_{2,1} & 0 & \cdots & C_{2,n} \\
\vdots & \vdots & \ddots & \vdots \\
C_{n,1} & C_{n,2} & \cdots & 0
\end{bmatrix},
\]
where \( C_{ij} \) denotes the travel cost from region \( A_i \) to region \( A_j \) (and \( C_{ii} = 0 \)). This cost is derived from the activated skeletal points associated with each region: each region \( A_i \) has a set of activated skeletal points \( V(A_i) \), and the cost between two regions is defined as the minimum skeletal distance between any pair of points across the two regions:
\begin{equation}
  C_{ij} = \min_{v_p \in V(A_i), v_q \in V(A_j)} d_{\text{skel}}(v_p, v_q),
\end{equation}
where $d_{\text{skel}}(v_p, v_q) $ denotes the skeletal graph distance, as denoted in Section \ref{prepro}. It is worth noting that while kinematic constraints (e.g., current velocity) typically render trajectory costs asymmetric in local planning, our framework decouples global guidance from local execution. In the global region-sequence planning phase, targets are spatially distant, making the traversal cost dominated by the skeletal path length rather than instantaneous dynamics. Therefore, given the undirected nature of the skeletal graph, the inter-region cost is inherently symmetric ($C_{ij} = C_{ji}$), and the problem is formulated as a Symmetric TSP with a fixed start node (Line~\ref{line:tsp}).

\begin{figure}[t]
    \centering
    \includegraphics[width=8.0cm]{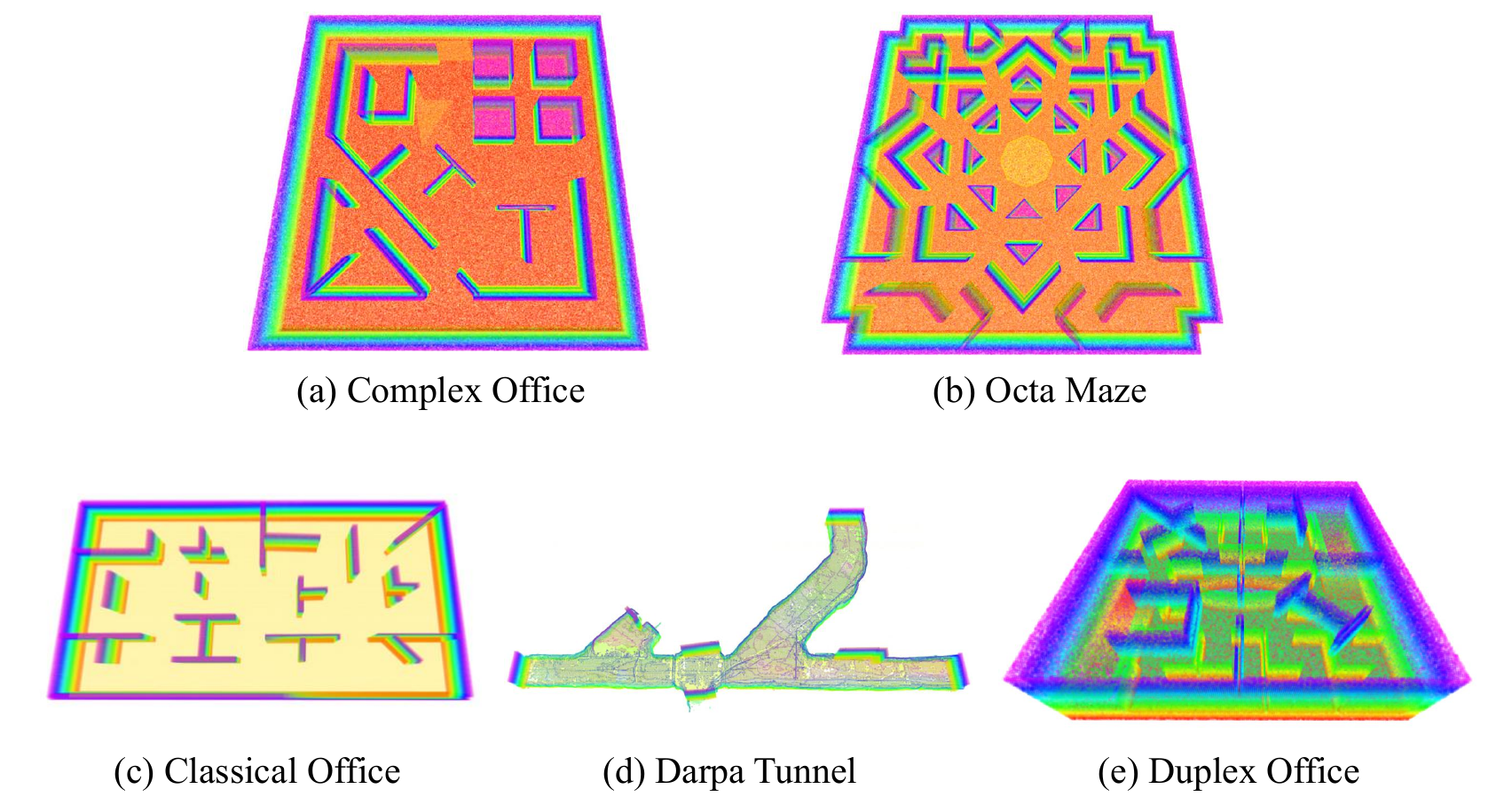}
    \caption{Simulation environments.}
    \label{environments}
\end{figure}

\section{Numerical Simulations}

\begin{table*}[t]
\centering
\caption{{Comparative Results of All Methods}}
\label{tab:exploration_stats}
\begin{tabular}{l l c c c c c c >{}c >{}c c c}
\toprule
Environment & Method & \multicolumn{2}{c}{Exploration Time (s)} & \multicolumn{2}{c}{Computational Cost ($ms$)} & \multicolumn{2}{c}{Flight Speed ($m/s$)}  & \multicolumn{2}{c}{Path Length ($m$)} & \multicolumn{2}{c}{Coverage ($m^3$)} \\
\cmidrule(lr){3-4} \cmidrule(lr){5-6} \cmidrule(lr){7-8} \cmidrule(lr){9-10} \cmidrule(lr){11-12}
& & Avg & Std & Avg & Std & Avg & Std & Avg & Std & Avg & Std\\
\midrule
\multirow{5}{*}{Complex Office} 
& FUEL & 277.31 & 16.17 & 34.15 & 4.19 & 1.35 & 0.03 & 375.02 & 29.34 & 1730.14 & 9.35\\
& FAEP & 208.93 & 7.02 & 39.46 & 4.51 & 1.32 & 0.05 & \textbf{275.98} & 12.96 & \textbf{1738.12} & 7.18 \\
& RACER & 252.33 & 21.64 & 127.62 & 16.29 & 1.60 & 0.02 & 405.06 & 48.13 & 1721.79 & 8.49 \\
& FALCON & 171.97 & 7.28 & 147.96 & 20.18 & 1.65 & 0.03 & 284.82 & 18.43  & 1734.12 & 6.89\\
& Proposed & \textbf{158.24} & 9.78 & \textbf{7.71} & 0.73 & \textbf{1.88} & 0.03 & 297.92 & 37.62 & 1727.61 & 9.28 \\
\midrule
\multirow{5}{*}{Octa Maze} 
& FUEL & 346.27 & 19.96 & 63.93 & 10.69 & 1.29 & 0.04 & 447.32 & 30.18 & 2059.74 & 8.25\\
& FAEP & 331.39 & 12.23 & 60.96 & 6.89 & 1.28 & 0.04 & 424.84 & 20.29 & 2043.24 & 6.79 \\
& RACER & \makecell{-} & \makecell{-} & \makecell{-} & \makecell{-} & \makecell{-} & \makecell{-} & \makecell{-} & \makecell{-} & \makecell{-} & \makecell{-}\\
& FALCON & 233.78 & 9.87 & 140.98 & 15.67 & 1.76 & 0.05  & \textbf{411.93} & 16.34 & 2058.08 & 7.06\\
& Proposed & \textbf{228.79} & 12.96 & \textbf{5.70} & 0.27 & \textbf{1.90} & 0.06 & 435.87 & 27.61 & \textbf{2069.33} & 9.89 \\
\midrule
\multirow{5}{*}{Classical Office} 
& FUEL & 150.14 & 9.67 & 19.87 & 3.14 & 1.24 & 0.03 & 187.04 & 15.07 & 881.76 & 4.97 \\
& FAEP & 141.78 & 7.01 & 25.21 & 3.98 & 1.22 & 0.03 & 173.98 & 10.41 & 870.79 & 6.24\\
& RACER & 124.14 & 3.96 & 41.17 & 4.98 & 1.54 & 0.05 & 192.21 & 8.79  & 884.17 & 3.94\\
& FALCON & \textbf{97.96} & 3.19 & 27.58 & 1.39 & 1.70 & 0.05 & \textbf{166.98} & 6.98 & \textbf{890.71} & 8.38 \\
& Proposed & 102.27 & 5.69 & \textbf{3.31} & 0.19 & \textbf{1.77} & 0.04 & 181.64 & 13.68 & 876.75 & 6.89\\
\midrule
\multirow{5}{*}{Darpa Tunnel} 
& FUEL & 63.48 & 3.93 & 37.14 & 3.20 & 1.40 & 0.04 & 89.34 & 6.98 & 488.37 & 6.87\\
& FAEP & 60.25 & 2.50 & 32.90 & 2.89 & 1.39 & 0.05 & \textbf{83.93} & 5.84 & 490.24 & 7.58\\
& RACER & 60.98 & 4.96 & 68.38 & 2.99 & 1.63 & 0.03 & 100.28 & 10.36 & 486.74 & 5.91\\
& FALCON & 50.16 & 2.91 & 48.59 & 3.96 & 1.89 & 0.04 & 95.23 & 4.01 & \textbf{493.12} & 5.89 \\
& Proposed & \textbf{47.36} & 2.98 & \textbf{3.51} & 0.14 & \textbf{1.95} & 0.05 & 92.91 & 6.24 & 487.15 & 6.10 \\
\midrule
\multirow{5}{*}{Duplex Office} 
& FUEL & 236.97 & 11.24 & 48.69 & 2.10 & 1.41 & 0.05 & 333.23 & 17.90 & 1464.01 & 5.57\\
& FAEP & 220.48 & 7.80 & 44.59 & 3.54 & 1.38 & 0.03 & 305.07 & 12.69 & 1477.36 & 8.12\\
& RACER & 243.01 & 10.87 & 61.24 & 6.30 & 1.59 & 0.05 & 387.20 & 16.24 & 1466.68 & 5.06\\
& FALCON & \textbf{176.92} & 5.21 & 78.11 & 7.90 & 1.72 & 0.05 & \textbf{304.99} & 10.01 & 1482.39 & 6.19 \\
& Proposed & 184.36 & 8.37 & \textbf{5.32} & 0.18 & \textbf{1.75} & 0.06 & 323.38 & 14.68 & \textbf{1488.31} & 7.17 
\\

\bottomrule
\end{tabular}
\end{table*}

\begin{figure*}[t]
    \center
    \includegraphics[width=17.5cm]{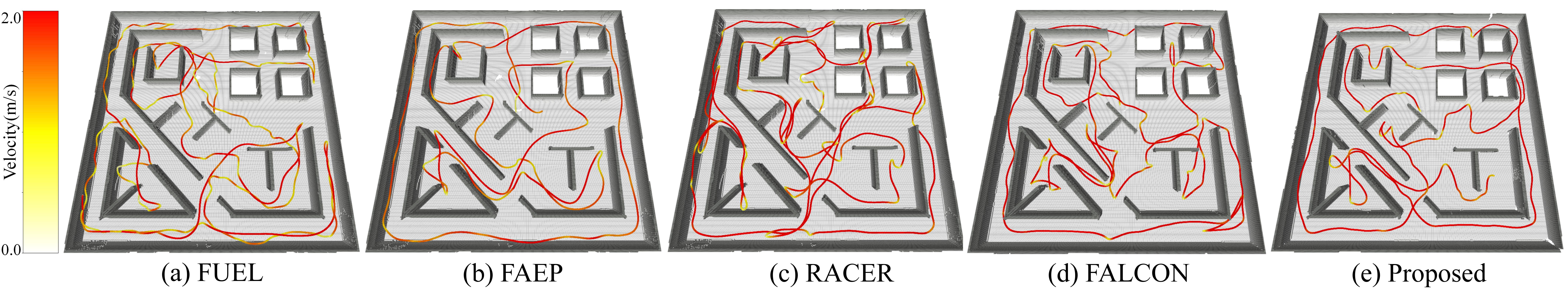}
    \caption{Visualization of exploration paths and velocity for four algorithms in the Complex Office environment, corresponding respectively to (a) FUEL, (b) FAEP, (c) RACER, (d) FALCON, (e) Proposed.}
    \label{Simulation}
\end{figure*}

To evaluate the proposed method SCOPE, this section compares it with state-of-the-art approaches through simulation experiments. Ablation studies are conducted to demonstrate the necessity of the various modules in SCOPE.

\subsection{Comparative Results}
Our simulations run on Ubuntu 20.04 with ROS implementation. The computing hardware consists of an Intel Core i5-9500H processor, RTX1650 4G graphics card, and 8G memory. The RGB-D camera used in simulations has a field of view of $[115^\circ, 92^\circ]$ and a maximum sensing range of $5.0  \mathrm{m}$. All algorithms are evaluated under identical motion constraints, where the linear velocity is constrained axis-wise to $2.0\,\text{m/s}$ (i.e., $v_x, v_y, v_z \in [-2.0, 2.0]\,\text{m/s}$) and the maximum angular velocity is set to $1.57\,\text{rad/s}$. We compare SCOPE with four baselines (FUEL~\cite{zhou2021fuel}, FAEP~\cite{zhao2023autonomous}, RACER~\cite{zhou2023racer}, FALCON~\cite{10816079}) within the five simulation environments (Complex Office, Octa Maze, Classical Office, Darpa Tunnel, Duplex Office, see Fig.~\ref{environments}). To ensure a fair comparison, we employed identical map bounding box settings for all algorithms in each environment, with the boundaries defined to be slightly larger than the actual map dimensions to accommodate boundary exploration. Each method was run ten times in the simulation, measuring exploration time, computational cost, UAV speed, path length and coverage.
To ensure fairness, benchmark experiments measure computation time in a unified way, excluding frontier computation as in FALCON.

Table~\ref{tab:exploration_stats} demonstrates that SCOPE achieves competitive exploration efficiency comparable to state-of-the-art methods, while offering a substantial reduction in computational cost.

In the Complex Office, SCOPE exhibits performance on par with the strongest baselines. Specifically, it reduces exploration time by 7.98\% compared to FALCON while slashing computational cost by 77.4\% compared to FUEL. As shown in Fig.~\ref{Simulation}, through the coordination of the proximal planner and the region-sequence planner, SCOPE reduces backtracking while maintaining trajectory smoothness. In contrast, FUEL exhibits oscillating motion near corridor edges by appending newly discovered residuals to the ATSP tail, forcing costly returns. Meanwhile, FAEP is limited by slow spiral paths, and RACER and FALCON suffer from void-region disruptions or frequent decelerations.

In the large-scale Octa Maze, our approach achieved the shortest time and lowest computational cost by reducing unnecessary global planning. Conversely, RACER failed as it suffered from topological oscillations and entrapment in local minima, likely exacerbated by its sensitivity to bounding box definitions and excessive computational demands, while FALCON suffered high computational overhead.
In small, structured maps (e.g., Classical Office), low POMDP uncertainty allows ATSP-based methods like FALCON to reach theoretical limits without oscillation. While SCOPE's efficiency is slightly lower here, it operates at only 7.22\% of the computational cost, while maintaining a coverage of 91.3\% of the areas, demonstrating superior cost-effectiveness by preserving resources for other high-level tasks.
In the DARPA Tunnel, SCOPE achieved the fastest exploration by maintaining high speed, though FALCON achieved marginally higher coverage.
In the Duplex Office, a dual-layer environment connected by stairs with full 3D topological complexity, FALCON achieves the shortest time (176.92\,s) and path length (304.99\,m) by searching for geometrically optimal frontiers across the entire volumetric space. However, this global optimality incurs a high cost of 78.11\,ms per cycle. In contrast, SCOPE adopts a hierarchical strategy. Although this abstraction introduces minor path sub-optimality ($\approx$ 6\% longer trajectory), it drastically reduces the computational load by 93.2\%, maintaining a lightweight footprint essential for limited onboard resources.

\begin{figure}[t]
    \centering
    \includegraphics[width=8.0cm]{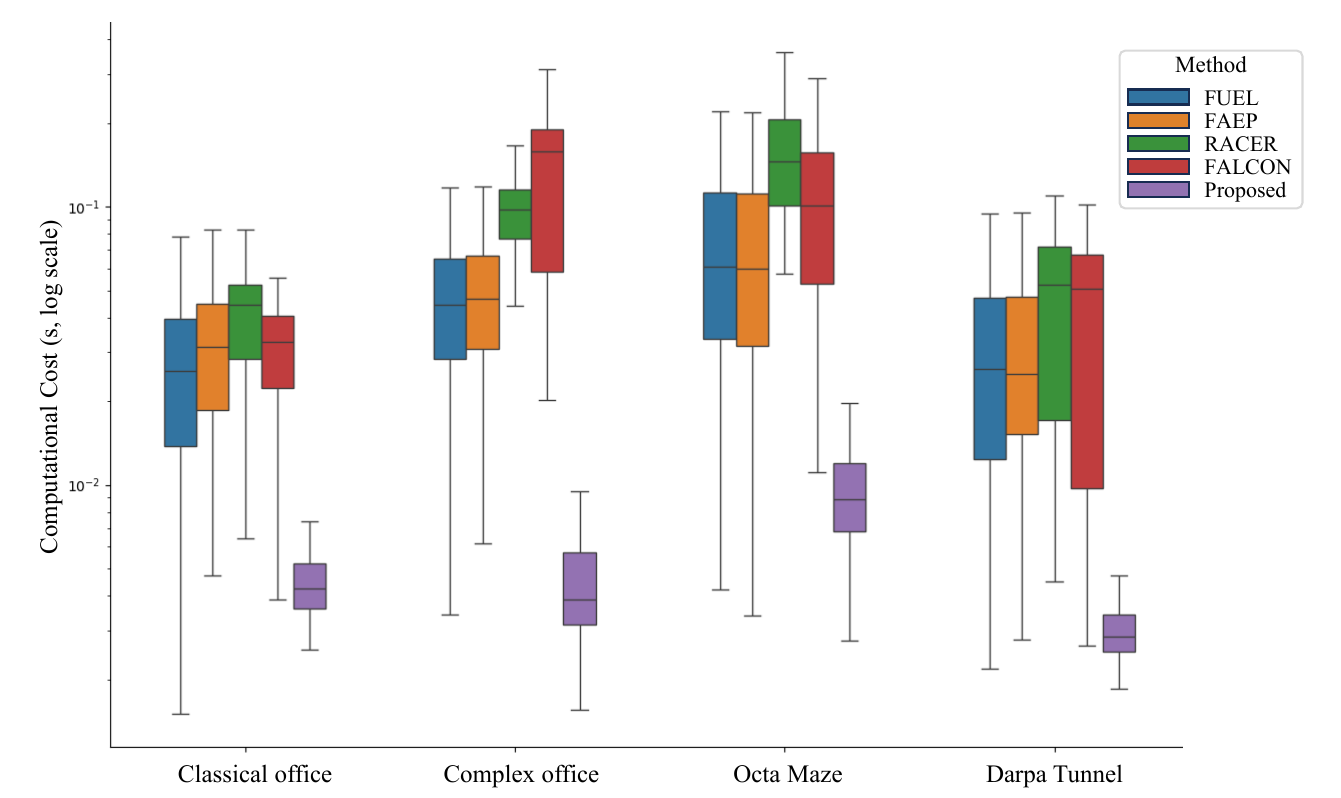}
    \caption{Boxplot of computation time per planning iteration (log scale).}
    \label{computation}
\end{figure}

In terms of computational cost, as shown in Fig.~\ref{computation}, the costs of FUEL and FAEP grow with frontier number, while RACER and FALCON incur higher overhead due to explicit decomposition and frequent global planning on volatile targets (optimizing a ``phantom future''). In contrast, SCOPE employs implicit unknown-space analysis and engages global guidance only when necessary. This strategy effectively amortizes the heavy optimization load, achieving significantly lower computational cost.

Overall, SCOPE achieves shorter exploration time and higher flight speed while maintaining comparable coverage. By decoupling local execution from global planning, our method mitigates the ``topological oscillation'' seen in baselines (where minor map updates trigger stop-and-go maneuvers). Although this hierarchical strategy trades minor geometric optimality (i.e., slightly longer paths) for system latency, it ensures continuous high-speed motion and reduces the computational cost by an average of 86.9\% compared with baselines, validating the efficiency of our motion-aware framework.

\subsection{Ablation Studies}

\begin{table}
\centering
\caption{Ablation Study Results Compared to the Initial Model}
\label{tab:ablation}
\begin{tabular}{ccc}
\hline
\textbf{Method Variant} & \textbf{Exploration Time ($s$)} & \textbf{Computational Cost ($ms$)}\\
\hline
w/o PIR   & 236.28 & 5.76\\
w/o RSP   & 269.18 & 5.15\\
w/o PP    & 261.89 & 22.98\\
w/o SG   & {225.17} & 43.67\\
Proposed     & 228.79 & 5.70\\
\hline
\end{tabular}
\end{table}

\begin{table}
\centering
\caption{Breakdown of Computational Time per Control Cycle}
\label{tab:comp_breakdown}
\begin{tabular}{l c c}
\toprule
\textbf{Module} & \textbf{Avg. Time ($ms$)} & \textbf{Percentage (\%)} \\
\midrule
Skeleton Update & 3.06 & 53.7\% \\
Region Analysis & 1.08 & 18.9\% \\
Planner(RSP/PP) & 0.57 (14.09 / 0.46) & 10.0\% \\
Trajectory Optimization & 0.99 & 17.4\% \\
\midrule
\textbf{Total System Latency} & \textbf{{5.70}} & \textbf{100.0\%} \\
\bottomrule
\end{tabular}
\end{table}

Ablation studies evaluated the contribution of four components: Prioritizing Isolated Regions (PIR), Region-Sequence Planner (RSP), Proximal Planner (PP), and the Skeletal Graph (SG, substituted with a Generalized Voronoi Diagram). In the absence of RSP, the system defaulted to greedy nearest-target selection, whereas removing PP forced RSP execution at every step. Results from 10 trials in the Octa Maze are summarized in Table~\ref{tab:ablation}.

Disabling PIR increased exploration time by 3.3\% due to less optimized pathing, with negligible computational impact. Removing RSP degraded efficiency by 17.7\% but only saved 9.6\% in computation, reflecting its low-frequency overhead. Conversely, removing PP increased exploration time by 14.5\% while causing a massive 303.2\% surge in computational cost, as frequent global replanning introduced unnecessary latency without strategic gain. Substituting the proposed skeleton with a standard Voronoi diagram yielded negligible exploration benefits but triggered a 666.1\% increase in computation, confirming that the heavy overhead of full Voronoi updates offers no advantage over our lightweight representation.

Table~\ref{tab:comp_breakdown} further details the computational distribution. Skeletal maintenance consumes the largest share (53.7\%) yet remains under 3.1\,ms, while the core planner averages just 0.57\,ms. This efficiency stems from our interleaved strategy: the heavier RSP ($\approx$14\,ms) is triggered intermittently to guide high-level decisions, while the lightweight PP runs frequently. This effectively amortizes the global optimization load, ensuring system responsiveness without bottlenecks.

\section{REAL-WORLD EXPERIMENTS}

\begin{figure}[t]
    \centering
    \includegraphics[width=7.0cm]{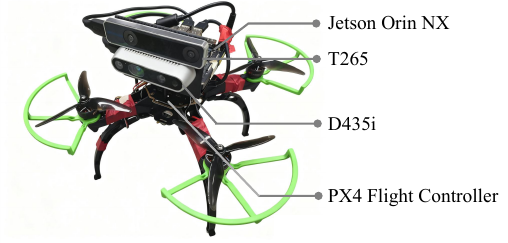}
    \caption{The UAV platform used in real-world experiments.}
    \label{uav}
\end{figure}

We evaluated SCOPE in real-world scenarios to verify the feasibility of our method for achieving autonomous exploration in practical applications. The UAV platform (Fig.~\ref{uav}) was equipped with a Jetson Orin NX 16GB as the onboard computing unit. The real-time pose was obtained using an Intel T265, depth information was acquired from an Intel D435i, and flight was controlled by a PX4 Flight Controller.

\begin{figure}[t]
    \centering
    \includegraphics[width=8.0cm]{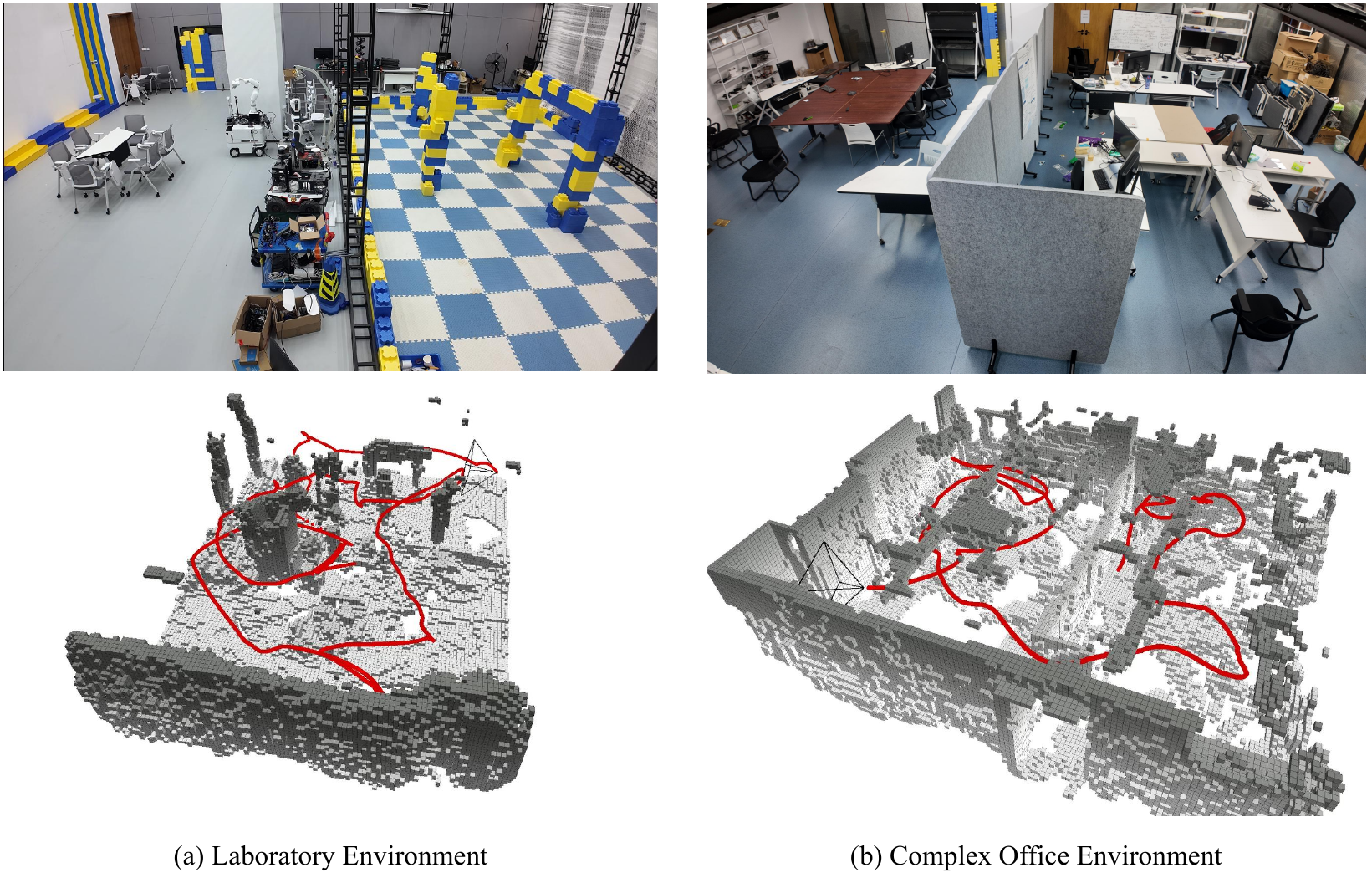}
    \caption{Exploration results in real-world environments: (a) is a laboratory with irregular obstacles. (b) is a cluttered office. The occupancy maps obtained in each case are displayed at the bottom.}
    \label{realworld}
\end{figure}

The UAV's kinematic limits were enforced via axis-wise constraints, with the maximum linear velocity $v_{\max} = 0.5\,\text{m/s}$, linear acceleration $a_{\max} = 0.5\,\text{m/s}^2$, and angular limits $\omega_{\max} = \alpha_{\max} = 1.05\,\text{rad}/\text{s}^{2}$. Two representative environments (Fig.~\ref{realworld}) were used to validate the framework's robustness and efficiency in real-world settings.

The first environment (Fig.~\ref{realworld}(a)) was an unstructured $13 \times 15 \times 2.5\,\text{m}^3$ laboratory containing irregular obstacles and complex occlusions. In this scenario, the region-sequence planner effectively reduced global planning dimensionality through adaptive node clustering. The UAV achieved a coverage of $421.31\,\text{m}^3$ in $96.12\,\text{s}$ with a minimal average computation cost of $5.81\,\text{ms}$, demonstrating high scalability in complex 3D layouts. The second environment (Fig.~\ref{realworld}(b)) was a $13 \times 12 \times 2.5\,\text{m}^3$ cluttered office space with dense furniture and frequent dead-ends. By leveraging the isolation score to prioritize these hard-to-reach regions, the planner minimized redundant backtracking and ensured smooth navigation. In this setting, the UAV achieved an exploration time of $78.83\,\text{s}$ ($323.50\,\text{m}^3$ coverage) and an average computation cost of $5.93\,\text{ms}$, confirming the method's effectiveness in constrained, high-entropy spaces.

Overall, the framework consistently maintained low computational overhead despite sensor noise and environmental clutter. These results highlight SCOPE's suitability for real-time autonomous exploration on resource-constrained embedded platforms.

\section{CONCLUSIONS}
In this letter, we have proposed SCOPE, a computationally efficient and motion-aware autonomous exploration framework. By maintaining a lightweight skeletal graph and decoupling local kinematic execution from global topological guidance, SCOPE ensures continuous high-speed motion with minimal planning overhead. Extensive experiments have demonstrated that our approach achieves exploration efficiency comparable to state-of-the-art methods while reducing the average computational cost by 86.9\%, proving its suitability for resource-constrained MAVs.

Future work will focus on three key directions: (1) developing adaptive triggering mechanisms that adjust the activation of the Region-Sequence Planner based on environmental connectivity and information gain; (2) integrating Deep Reinforcement Learning to enhance decision-making in complex environments; (3) extending the system to LiDAR-based platforms to enable robust deployment in large-scale, unstructured outdoor scenarios.

\bibliographystyle{IEEEtran}
\bibliography{IEEEabrv, references}

\end{document}